\documentclass[conference]{IEEEtran}
\IEEEoverridecommandlockouts

\usepackage{cite}
\usepackage{amsmath,amssymb,amsfonts}
\usepackage{graphicx}
\usepackage{textcomp}
\usepackage{xcolor}
\usepackage{amsmath,amsfonts}
\usepackage{algorithm}
\usepackage{array}
\usepackage[caption=false,font=normalsize,labelfont=sf,textfont=sf]{subfig}
\usepackage{textcomp}
\usepackage{utfsym}
\usepackage{pifont}
\usepackage{stfloats}
\usepackage{url}
\usepackage{verbatim}
\usepackage{hyperref}

\hypersetup{
	colorlinks=true,
	linkcolor=[rgb]{1,0.058,0.121},
	filecolor=cyan,  
	urlcolor=[rgb]{0.948,0.008,0.56},
	citecolor=[rgb]{0.298,1,0.309},
}

\usepackage{makecell}
\usepackage{bm}

\definecolor{LightCyan}{rgb}{0.88,1,1}

\definecolor{sgreen}{RGB}{30, 150, 30}

\usepackage{amsthm,amsmath,amssymb}
\usepackage{mathrsfs}
\usepackage{booktabs}
\usepackage{multicol}
\usepackage{color}
\usepackage{float}
\usepackage{subfig}

\usepackage{enumitem}
\usepackage{bbding}

\usepackage{float}
\usepackage{subfig}
\usepackage{colortbl}
\usepackage{rotating}
\usepackage{array}

\definecolor{LightCyan}{rgb}{0.88,1,1}
\definecolor{cvprblue}{rgb}{0.21,0.49,0.74}

\usepackage{adjustbox}

\usepackage{tabularx}
\usepackage{etoolbox}
\usepackage{makecell}
\usepackage{pifont}

\usepackage{wrapfig}
\usepackage{cutwin}
\usepackage{alltt}

\usepackage{spconf,epsfig}
\usepackage{multirow} 
\usepackage{algpseudocode}

\newcommand{\mytitlefont}{\fontsize{18.5}{21}\selectfont}

\def\BibTeX{{\rm B\kern-.05em{\sc i\kern-.025em b}\kern-.08em
    T\kern-.1667em\lower.7ex\hbox{E}\kern-.125emX}}

\begin{document}

\title{\mytitlefont 3A-YOLO: New Real-Time Object Detectors with Triple Discriminative Awareness and Coordinated Representations}

\name{Xuecheng Wu$^{1\dagger}$, Junxiao Xue$^{2\ast\dagger}$, Liangyu Fu$^{3}$, Jiayu Nie$^{1}$, Danlei Huang$^{1}$, Xinyi Yin$^{4}$\thanks{$^{\ast}$Corresponding author. $^{\dagger}$Equal contributions.}}
\address{$^{1}$School of Computer Science and Technology, Xi'an Jiaotong University, Xi'an, China\\$^{2}$Research Center for Space Computing System, Zhejiang Lab, Hangzhou, China\\$^{3}$School of Software, Northwestern Polytechnical University, Xi'an, China \\$^{4}$School of Cyber Science and Engineering, Zhengzhou University, Zhengzhou, China\\ \tt xuecwu@gmail.com, xuejx@zhejianglab.cn}

\maketitle

\begin{abstract}
Recent research on real-time object detectors (\textit{e.g.}, YOLO series) has demonstrated the effectiveness of attention mechanisms for elevating model performance. Nevertheless, existing methods neglect to unifiedly deploy hierarchical attention mechanisms to construct a more discriminative YOLO head which is enriched with more useful intermediate features. To tackle this gap, this work aims to leverage multiple attention mechanisms to hierarchically enhance the triple discriminative awareness of the YOLO detection head and complementarily learn the coordinated intermediate representations, resulting in a new series detectors denoted 3A-YOLO. Specifically, we first propose a new head denoted TDA-YOLO Module, which unifiedly enhance the representations learning of scale-awareness, spatial-awareness, and task-awareness. Secondly, we steer the intermediate features to coordinately learn the inter-channel relationships and precise positional information. Finally, we perform neck network improvements followed by introducing various tricks to boost the adaptability of 3A-YOLO. Extensive experiments across COCO and VOC benchmarks indicate the effectiveness of our detectors.
\end{abstract}


\section{Introduction}
\label{sec:intro}
Real-time object detection aims at answering ``what objects are located where" for the resource-constrained application scenarios. Recently, the increasing adoption of deep neural networks (DNNs) has significantly facilitated the developments of real-time object detection, introducing a plethora of new ideas and methods to this research field~\cite{b11,b3}.

The current mainstream methods can be primarily divided into one-stage detectors and two-stage detectors. In recent years, Faster R-CNN~\cite{b11} and many other methods, which are based on region proposals, have been introduced. Later, relevant researchers propose various one-stage detectors, such as SSD~\cite{b8}, YOLOv3~\cite{b7}, and EfficientDet~\cite{b12}. Among many one-stage detectors, the YOLO series, such as \cite{b1,b3,b7,YOLOv7}, is one of the most famous series detectors. The YOLO family always pursues the best speed and accuracy balance for extensive computer vision applications. YOLOv3~\cite{b7} has primarily changed the network paradigm of the YOLO series detectors. YOLOv4 deploys a variety of augmentation strategies, such as ``bag of freebies" and ``bag of specials", which help to improve model performance. As for YOLOv5~\cite{b3}, researchers prefer to view it as an industrial version of YOLOv4. Regarding more advanced YOLO detectors such as YOLOv7~\cite{YOLOv7}, the community often regards them as the ensemble methods, which incorporate numerous specialized designs.

\begin{figure}[t]
\centering
\includegraphics[height=5.5cm, width=\linewidth]{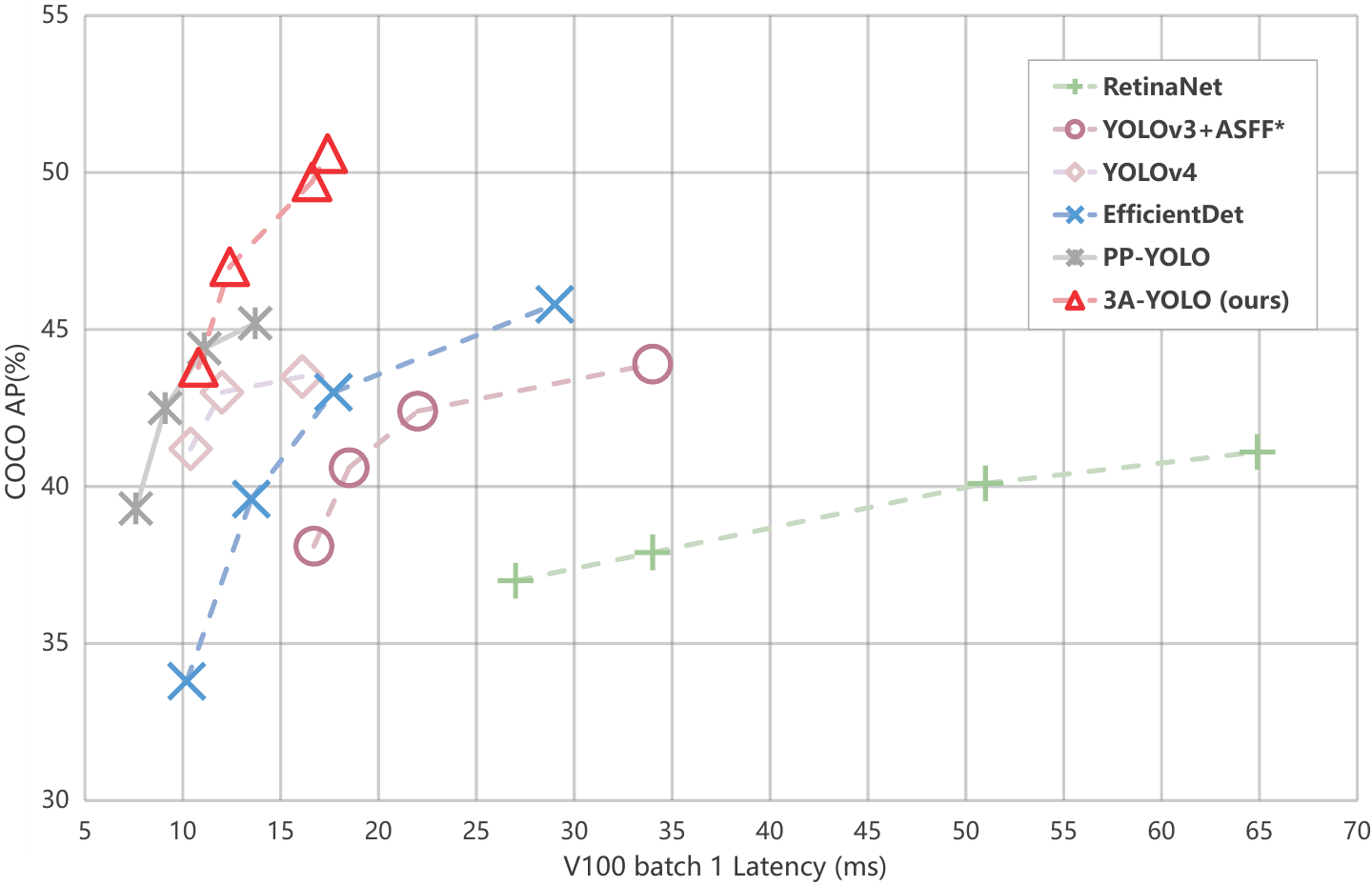}
\vspace{-1em}
\caption{Speed and accuracy trade-off of our 3A-YOLO series detectors and representive CNNs-based real-time detectors on COCO dataset. For example, our 3A-YOLO with 608 $\times$ 608 resolution increases YOLOv4~\cite{b1} by 6.2\% AP and maintains a competitive speed of 60.1 FPS.}
\label{fig:0}
\vspace{-1.0em}
\end{figure}

Meanwhile, in recent years, studies related to attention mechanisms in computer vision have also emerged significantly~\cite{b13,b53}. Furthermore, attention mechanisms have been extensively proven to be beneficial in real-time object detection. Recent studies~\cite{EBSE-YOLO,YOLO-SE,SBMYv3_Samal} have extensively introduced the soft attention mechanisms (\textit{e.g.}, SE~\cite{b13}, BAM~\cite{b14}, CBAM~\cite{b17}, and ECA~\cite{b53}) into the YOLO series detectors. Specifically, YOLO-SE~\cite{YOLO-SE} designs three attention-related modules to better perform the multi-scale and samll-object detection. Samal et al.~\cite{SBMYv3_Samal} introduce BAM~\cite{b14} to construct S3Pooling for automatic detection of obscene content. EBSE-YOLO~\cite{EBSE-YOLO} deploys the ECA-Net~\cite{b53} to help model focus more on the small targeted objects. Chung et al.~\cite{YOLO-SLD_Chung} present an improved YOLOv7 integrating the parameter-free attention module for license plate detection. Despite promising results, the above methods neglect to unifiedly deploy the hierarchical attention mechanisms to construct a comprehensive YOLO head consolidated with more discriminative representations. At the same time, recent work~\cite{b2} has demonstrated that attention mechanisms are essential to building a high-performance detection head that should have the capability of triple discriminative awareness (\textit{i.e.}, scale-awareness, spatial-awareness, and task-awareness). Furthermore, the coordinated representations of inter-channel and positional information is also important for a high-performance detection head, which facilitates to improve the capability of multi-scale perception, spatial localization, as well as the feature differentiation. However, the popular soft attention components~\cite{b13,b14,b17} still have limitations in coordinately learning representations. SE~\cite{b13} only considers the feature encoding between channels but ignores the significance of positional information, which is crucial for capturing various structures in object detection. Later works like BAM~\cite{b14} and CBAM~\cite{b17} attempt to make use of precise positional information by reducing channel dimensions and using convolutions for spatial attention computation. However, convolutions can only model local relevance, failing to capture the long-range dependencies.

To this end, this work targets at leveraging multiple attention mechanisms to unifiedly enhance the triple discriminative awareness of the detection head based on the vanilla YOLO baseline~\cite{b1} and complementarily learn the coordinated intermediate representations, leading to a new series detectors with different input resolutions denoted 3A-YOLO. In detail, we hierarchically introduce triple discriminative attention components on the designated dimensions of features ($i.e.$, scale-wise, spatial-wise, and channel-wise), resulting in an intensively triple-awareness enhanced YOLO head termed TDA-YOLO Module. Subsequently, we steer the intermediate features to learn the inter-channel relationships and precise positional information in a coordinated manner, thereby complementarily enhancing the adaptability of multi-scale perception and percise spatial localization for our new detection head. In the end, we emphatically perform structure enhancements of the neck network and introduce various tricks to further boost the robustness and generalization of the overall detectors. Based on above improvement measures, we extraly propose three scaled versions with different model capacities (\textit{i.e.}, 3A-YOLO-X, 3A-YOLO-Tiny, and 3A-YOLO-Nano) to meet the diverse needs of varying real-world scenarios. To demonstrate the effectiveness of our methods, we perform extensive experiments on the popular COCO~\cite{b15} and VOC~\cite{b48} datasets. As illustrated in Fig.~\ref{fig:0}, our detectors exhibit superior speed and accuracy trade-off compared with representive CNNs-based real-time detectors. Moreover, ablation studies also justify various design choices in our models.

The contributions of this paper are three-fold: \textbf{(1)} We present a new series real-time detectors denoted 3A-YOLO, which totally include three scaled capacities. We propose the TDA-YOLO Module, which unifiedly enhances triple discriminative awareness leveraging hierarchical attention components. \textbf{(2)} We steer the intermediate features to efficiently learn coordinated representions, and then introduce neck network improvements followed by various tricks for better adaptability. \textbf{(3)} Extensive experiments across COCO and VOC datasets demonstrate the effectiveness of our detectors. The ablations are also performed to explore the crucial factors of our design.

\section{Related Work}
\label{sec:related}
\noindent \textbf{Modern Real-time Object Detectors.}~Nowadays, most existing real-time detectors are under the anchor-based one-stage manner~\cite{b1,b3,b12,b16}. The modern one-stage detectors generally consist of two parts: a pre-trained backbone and precise detection heads utilized for localization and classification. Furthermore, some one-stage detectors plug in several intermediate layers (\textit{i.e.}, neck network) between two parts to progressively strengthen feature fusion and representation. First, there are various backbones, such as ResNet~\cite{b28} and DarkNet~\cite{b30}. Secondly, a neck network comprises multiple top-down and bottom-up paths, primarily involving FPN~\cite{b58} and PAN~\cite{b36}. Finally, the detection heads mainly include YOLO Head~\cite{b1}, SSD~\cite{b8}, and RetinaNet~\cite{b16}. Despite achieving impressive performance within their respective frameworks, existing heads lack a unified capability of triple discriminative awareness, which are scale-awareness, spatial-awareness, and task-awareness, respectively. To tackle the gaps, this work proposes a unifiedly enhanced YOLO head (\textit{i.e.}, TDA-YOLO Module) based on three hierarchical attention components.

\noindent \textbf{Attention Mechanisms in Real-time Object Detection.}~Recently, researchers have extensively leveraged the attention mechanisms~\cite{b4,b13,b17} for enhancing real-time detection performance~\cite{b49}. Concretely, SE~\cite{b13} compresses features to establish inter-channel relationship at low computational costs. CBAM~\cite{b17} further extends the inspiration of SE by encoding positional information using convolutions with large kernels. Later works~\cite{b31,b34} further design various spatial components to capture cross-region information. Nevertheless, these works reveal a gap in coordinating the learned representations of channel-aware relationship and positional details. Distinct from previous methods, Coordinate Attention (CA)~\cite{b4} integrates positional information into channels, enabling the model to effectively capture channel-aware and position-sensitive features. Drawing on this insights, this work efficiently guides models to learn more coordinated intermediate representations for the improved input adaptability of our new detection head.

\begin{figure*}[t!]
\centering
\includegraphics[scale=0.53]{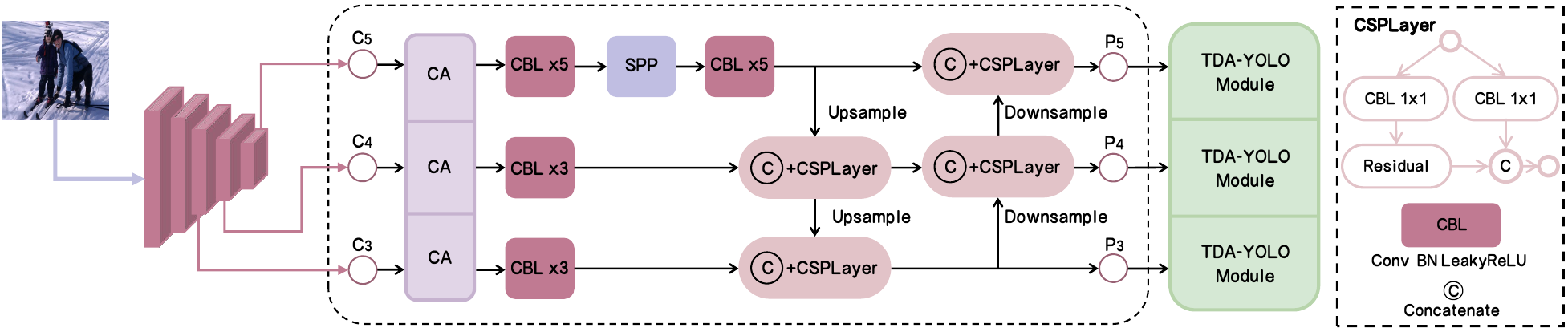}
\vspace{-0.80em}
\caption{The illustration of our 3A-YOLO. SPP is the Spatial Pyramid Pooling Layer~\cite{b41}. $C_i$ and $P_i$ ($i = \{3,4,5\}$) all denote the intermediate features.}
\vspace{-1.4em}
\label{fig:1}
\end{figure*}

\section{Methodology}
We utilize YOLOv4~\cite{b1} and YOLOv4-Tiny~\cite{b9} as the baselines due to their lower integration complexities, which allows for a more clear efficacy demonstration of our improvement measures. The illustration of our model is shown as Fig.~\ref{fig:1}.

\subsection{New Detection Head:~TDA-YOLO Module}
How to construct a high-performance detection head with triple discriminative awareness in the unified manner? A natural solution is to establish the whole self-attention mechanism across features. However, the optimization is difficult coupled with the unbearable computational overheads.

Building upon the insights from~\cite{b2}, we deploy hierarchical attention mechanisms on each designated dimension of feature (\textit{i.e.}, scale-wise, spatial-wise, and channel-wise) to unify triple discriminative awareness in a more efficient way. We thus propose the TDA-YOLO Module, including a ``Concat" layer, two adapted Dynamic Blocks~\cite{b2}, a ``Recover" layer, and the original YOLO Head, as displayed in Fig.~\ref{fig:A3E-YOLO} below. Given the intermediate output features $F_{i} \in \mathbb{R}^{C \times H \times W}, i=\{1,2,3\}$, we first stack $F_1$ with itself in the ``Concat" layer since the neck network has performed sufficient feature interactions to make up for the discrepancy of object scales contained in the feature maps for three different levels, leading to $F_1 \in \mathbb{R}^{S_{L} \times H \times W \times C}$, where $S_{L}$ represents the number of feature level, $H$, $W$, and $C$ denote the height, width, and channels, respectively. We then transform $F_1$ into $F \in \mathbb{R}^{S_{L} \times S \times C}$ by defining $S = (H \times W)$.

Subsequently, $F$ runs into two adapted Dynamic Blocks~\cite{b2} which are sequentially stacked. The overall formulation of one adapted Dynamic Block can be represented as:
\begin{equation}
W\left(F\right)=\xi_{C}\left(\xi_{S}\left(\xi_{S_{L}}\left(F\right) \cdot F\right) \cdot F\right) \cdot F,
\label{eq-1}
\end{equation}
where $\xi_{S_{L}}(\cdot)$, $\xi_{S}(\cdot)$, $\xi_{C}(\cdot)$ are scale-aware, spatial-aware, and task-aware attention components operating on the $S_{L}$, $S$, and $C$ dimensions, respectively. Next, we explain the discriminative role of each attention component as below. 

\noindent \textbf{Scale-aware Attention $\xi_{S_{L}}$}: Since the scale diversity is interrelated to features across levels, we introduce this component to dynamically learn the accessible relationship of the feature map based on its fused semantic importance, \textit{i.e.},
\begin{equation}
\xi_{S_{L}}\left(F\right) \cdot F=\sigma\left(\delta\left(\frac{1}{SC} \sum_{S, C} F\right)\right) \cdot F,
\label{math:2}
\end{equation}
where $\delta(\cdot)$ is a linear transformation approximated by one $1 \times 1$ convolution, and $\sigma(x)=\max \left(0, \min \left(1, \frac{x+1}{2}\right)\right)$ refers to a hard sigmoid function.

\noindent \textbf{Spatial-aware Attention $\xi_{S}$}: Considering that the geometric transformations of various object shapes are correlated to features at different spatial locations, we deploy $\xi_{S}$ based on the generated scale-sensitive features to focus on the regions that always co-exist among various spatial locations. Due to the high dimensionality of $S$, we factorize this component into two steps: (1) sparse attention learning via a deformable convolution transformation; (2) coherent information aggregation of the fused feature map. The process can be formulated as:
\begin{equation}
\xi_{S}\left(F\right) \cdot F=\sum_{k=1}^{K} W_{l=1, k} \cdot F\left(l=1 ; p_{k}+\Delta p_{k} ; c\right) \cdot \Delta m_{k},
\label{math:3}
\end{equation}
where $K$ refers to the number of sparse sampling locations, $p_{k}+\Delta p_{k}$ denotes a shifted location by the self-learned spatial offset $\Delta p_{k}$ to center on the specific region, and $\Delta m_{k}$ denotes a self-learned importance scalar at the location $p_{k}$~\cite{b2}.

\begin{figure}[t!]
\centering
\includegraphics[width=\linewidth]{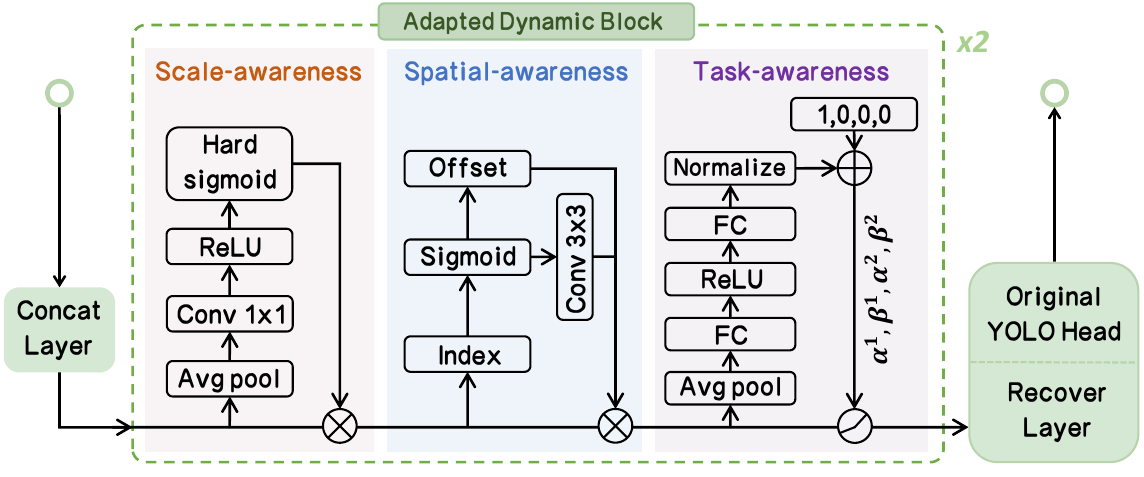}
\vspace{-1.7em}
\caption{The illustration of our proposed TDA-YOLO Module.}
\vspace{-1.5em}
\label{fig:A3E-YOLO}
\end{figure}

\noindent \textbf{Task-aware Attention $\xi_{C}$}: The task forms of one-stage detector are bounding box regression, classification, and confidence computation. The intrinsic feature distribution coupling between bounding box regression and classification impedes precise object localization and categorization. To achieve effective representations and generalization of various task forms in a coupling manner, we employ the adapted DY-ReLU-A~\cite{b39} to build $\xi_{C}$. The adapted DY-ReLU-A encodes the global information as a hyper function and accordingly adjusts the piecewise linear activation function. It can dynamically switch feature channels to support various task forms according to different convolutional kernel responses from objects, \textit{i.e.},
\begin{equation}
\scalebox{0.8605}{$\xi_{C}\left(F\right) \cdot F=\max \left(\alpha^{1}\left(F\right) \cdot F_{c}+\beta^{1}\left(F\right), \alpha^{2}\left(F\right) \cdot F_{c}+\beta^{2}\left(F\right)\right)$},
\label{math:4}
\end{equation}
where $F_{c}$ denotes the feature slice at the $c$-th feature channel and $\theta(\cdot)=\left[\alpha^{1}, \beta^{1}, \alpha^{2}, \beta^{2}\right]^{\mathrm{T}}$ is a hyper function which learns to variably control the activation thresholds~\cite{b2,b39}. $\theta(\cdot)$ first performs global average pooling along the $S_{L} \times S$ dimension, then sequentially goes through two fully connected layers, a ReLU layer, a normalization layer, and finally normalizes the output to [-1,1] utilizing a hard sigmoid function.

So far, the output features exhibit enhanced discriminative capabilities across multi-scale and spatial transformations, characterized by the aggregated information that uniformly improves the task-aware representations. Finally, we reshape the output as $F \in \mathbb{R} ^ {C \times H \times W}$ employing the ``Recover" layer, and then perform refined fusion and dimension adjustments through the original YOLO head. The same goes for the other features at different levels. In this way, we successfully deploy the unified attention components to enhance the capability of triple discriminative awareness in the detection head.

\subsection{Coordinated Representations Learning}
We steer the intermediate features to coordinately learn the channel-aware relationships and precise positional information via introducing the Coordinate Attention (CA)~\cite{b4}, thereby complementarily improving the crucial adaptability of multi-scale perception and geometric transformations for our TDA-YOLO Module. Specifically, the process of introduced CA can be decomposed into two stages: (I) coordinated information embedding; (II) coordinated attention generation.

In stage (I), given input $x_{c}(i, j)$, we deploy two spatial extents of pooling kernels $(H,1)$ and $(1,W)$ to perform a pair of 1D feature information encoding along the horizontal and vertical directions, leading to the outputs $q_{c}^{h}$ and $q_{c}^{w}$ of $c$-th feature channel at height $h$ and width $w$, \textit{i.e.},
\begin{align}
q_{c}^{h}(h) &= \frac{1}{W} \sum_{0 \leq i<W} x_{c}(h, i),\\
q_{c}^{w}(w) &= \frac{1}{H} \sum_{0 \leq j<H} x_{c}(j, w).
\label{math:5-6}
\end{align}

The above two transformations aggregately generate two direction-aware feature maps $q^{h}$ and $q^{w}$, promoting to capture the remote spatial interactions with refined positional details.

During stage (II), we first stack the globally aggregated $q^{h}$ and $q^{w}$, then compress channels by one $1 \times 1$ convolution. Next, we encode the precise positional information in both horizontal and vertical directions through a BatchNorm layer and a ReLU layer. Afterwards, $f$ is divided into $f^h$ and $f^w$ along the spatial dimensions, and two convolutions are utilized to transform $f^h$ and $f^w$ to features with the same dimensions of input $x_{c}(i, j)$. Finally, we output $g^h$ and $g^w$ via normalized weighting. The above process can be formulated as: 
\begin{align}
g^{h} &= \sigma\left(F_{h}\left(f^{h}\right)\right), \\
g^{w} &= \sigma\left(F_{w}\left(f^{w}\right)\right), \\
f     &= \delta\left(F_{s}\left(\left[q^{h}, q^{w}\right]\right)\right),
\label{math:7-9}
\end{align}
where $F_{s}(\cdot)$ denotes stacking followed by one $1 \times 1$ convolution and a BatchNorm layer. $\delta(\cdot)$ represents a ReLU layer. $f \in \mathbb{R} ^ {C / r \times (H + W)}$ is the encoded feature, and $r=16$ denotes the dimension reduction ratio for the scaling component size. $F_{h}(\cdot)$ and $F_{w}(\cdot)$ denote $1 \times 1$ convolution transformations, and $\sigma(\cdot)$ represents a sigmoid function. $g^h$ and $g^w$ are then extended as the attention weights, leading to the final output:
\begin{align}
y_{c}(i, j)=x_{c}(i, j) \times (g_{c}^{h}(i) \times g_{c}^{w}(j)),
\label{math:10}
\end{align}
where $x_c(i, j)$ refers to the input, $g_{c}^{h}(i)$ and $g_{c}^{w}(j)$ denotes the horizontal and vertical weights. $y_{c}(i, j)$ is the output feature.

As illustrated in Fig.~\ref{fig:1}, we first apply CA for $C3$ and $C4$ features to enhance the capture of small objects and encode positional details in long-range areas. We then deploy CA for $C5$ feature to deepen inter-regional correlations, thereby promoting the understanding of more comprehensive semantics.

\subsection{Neck Network Improvements \& Tricks Deployment}
We introduce a series of neck network improvements to better tackle the challenging detection environments and objects with similar appearances. We first expand the convolution layers around the SPP~\cite{b41} layer from three to five, then increase the convolution layers from one to three before $C3$ and $C4$ features. By replacing the five convolution blocks with CSPLayers, we further reduce computational costs while maintaining capacities. Experiments on COCO demonstrate a 0.4M reduction in parameters without compromising performance.

To further enhance the robustness and generalization of our detectors, we first employ Mosaic as the primary augmentation strategy, which simultaneously transforms four images through cropping, flipping, scaling, and RGB modifications. We then optimize the overall loss via the $\alpha$-balanced variant of Focal Loss~\cite{b16}, and apply label smoothing to mitigate overfitting. For non-maximum suppression, we adopt the DIoU-NMS~\cite{b26}, which extends traditional IoU by considering center point distances between bounding boxes, allowing more flexible IoU threshold selections. Besides, we explore the Dilated Encoder~\cite{b46} to expand receptive fields, though it is ultimately treated as an optional module due to its suboptimal efficiency.

\subsection{Three Scaled Versions of 3A-YOLO}
\noindent \textbf{3A-YOLO-X.}~To expand the capacity, we adopt the backbone of YOLOv5~\cite{b3} denoted Modified CSP v5. During model training, we deploy the input resolution of 640 $\times$ 640, and maintain almost the same learning strategy of~\cite{b3} to optimize parameters within this model. Furthermore, we employ stronger Mosaic augmentation strategy to adapt larger parameters.

\begin{figure}[t!]
\centering
\includegraphics[scale=0.65]{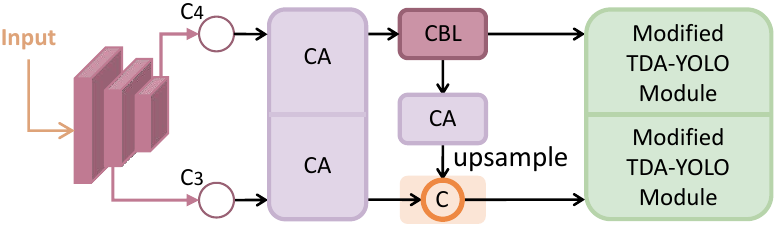}
\vspace{-0.8em}
\caption{The overall structure of our 3A-YOLO-Tiny.}
\vspace{-1.6em}
\label{fig:6}
\end{figure}

\noindent \textbf{Tiny and Nano Detectors.}~We propose 3A-YOLO-Tiny to meet the needs of various application scenarios~\cite{b9,b23}, as shown in Fig.~\ref{fig:6}. We first deploy the CSPDarknet53-Tiny~\cite{b30} as the backbone, and then utilize only one adapted Dynamic Block in the TDA-YOLO Module. Finally, we deploy CA~\cite{b4} for three times. For the mobile platforms, we utilize depth-wise convolutions to build 3A-YOLO-Nano. Besides, we employ the Mixup data augmentation strategy instead of Mosaic.

\begin{table*}[t!]
\centering
\caption{Accuracy and speed comparisons of our 3A-YOLO with representive real-time object detectors on MS-COCO~\emph{(test-dev 2017)}.}
\vspace{-0.5em}
\setlength{\arrayrulewidth}{0.37pt}
\renewcommand{\arraystretch}{1.03}
\resizebox{\linewidth}{!}{
\begin{tabular}{lcccccccccc}
\toprule
Method  & Backbone & Size & FPS & AP(\%) & $\mathrm{AP_{50}(\%)}$ & $\mathrm{AP_{75}(\%)}$ & $\mathrm{AP_{S}(\%)}$ & $\mathrm{AP_{M}(\%)}$ & $\mathrm{AP_{L}(\%)}$ \\
\hline
YOLOv3 + ASFF*~\cite{b37}     & Darknet-53        & 608           & 45.5               & 42.4            & 63.0          & 47.4          & 25.5         & 45.7         & 52.3         \\
RetinaNet~\cite{b16} & ResNet-101        & 1024          & 15.4               & 41.1            &  \textbf{--}            &         \textbf{--}     &     \textbf{--}        &    \textbf{--}         &     \textbf{--}        \\
EfficientDet-D2~\cite{b12} & Efficient-B3      & 896           & 34.5               & 45.8            & 65.0          & 49.3          & 26.6         & 49.4         & 59.8         \\
PP-YOLO~\cite{b23}     & ResNet50-vd-dcn   & 512           & 89.9               & 44.4            & 64.6          & 48.8          & 24.4         & 47.1         & 58.2         \\
YOLOv4~\cite{b1}   & CSPDarknet-53     & 608           & 62.0                 & 43.5            & 65.7          & 47.3          & 26.7         & 46.7         & 53.3         \\
YOLOv4-CSP~\cite{b9}  & Modified CSPDarknet53      & 640           & 73.0                 & 47.5            & 66.2          & 51.7          & 28.2         & 51.2         & 59.8         \\ 
YOLOv5-L~\cite{b3}          & Modified CSP v5   & 640           & 69.8               & 48.2
      & 66.9          & \textbf{--}             & \textbf{--}            & \textbf{--}            & \textbf{--}            \\ 
PP-YOLOv2~\cite{PP-YOLOv2} & ResNet101-vd-dcn      & 640          & 50.3               & 50.3            & 69.0          & 55.3          & 31.6         & 53.9        & 62.4         \\
\rowcolor{cvprblue!20}
\textbf{3A-YOLO (Ours)}          & CSPDarknet-53     & 416           & 92.6               & 43.8            & 64.1          & 48.1          & 29.5       & 49.6         & 54.2         \\
\rowcolor{cvprblue!20}
\textbf{3A-YOLO (Ours)}           & CSPDarknet-53     & 512           & 80.3                 & 47.0      & 65.6              & 51.4            & 30.4           & 51.5             & 58.3            \\
\rowcolor{cvprblue!20}
\textbf{3A-YOLO (Ours)}           & CSPDarknet-53     & 608           & 60.1             & 49.7        & 68.1              & 54.4              & 32.7             & 53.0             & 60.5            \\
\rowcolor{cvprblue!20}
\textbf{3A-YOLO-X (Ours)}           & Modified CSP v5  &640    & 57.4     & \textbf{50.6}    & 68.9      & 56.3     & 33.2      & 53.1     & 62.2 \\
\hline
\end{tabular}}
\label{tab:duibi-1}
\vspace{-1.8em}
\end{table*}

\begin{table}[t!]
\caption{Comparisons of 3A-YOLO with representive real-time object detectors in terms of $\mathrm{AP_{50}}$(\%) on PASCAL VOC~\emph{(test 2012)}.}
\centering
\vspace{-0.5em}
\setlength{\arrayrulewidth}{0.37pt}
\renewcommand{\arraystretch}{1.03}
\resizebox{\linewidth}{!}{
\begin{tabular}{lcccc}
\toprule
Method & Backbone & Size & $\mathrm{AP_{50}}$(\%) \\ \hline
YOLOv4~\cite{b1}           & CSPDarknet-53     & 416$\times$416       & 85.8              \\
HSD~\cite{b59}             & VGG16             & 512$\times$512       & 83.0              \\
BlitzNet~\cite{b50}        & ResNet-50         & 512$\times$512       & 81.5              \\
PS-KD~\cite{b51}           & ResNet-152        & 416$\times$416       & 79.7              \\
YOLOv5-L~\cite{b3}         & Modified CSP v5   & 640$\times$640       & 89.4              \\
CoupleNet~\cite{b52}       & ResNet-101        & 416$\times$416       & 82.7              \\
\rowcolor{cvprblue!20}
\textbf{3A-YOLO (Ours)}                   & CSPDarknet-53     & 608$\times$608       & \textbf{93.1}     \\ \hline
\end{tabular}}
\vspace{-1.0em}
\label{tab:duibi-2}
\end{table}

\section{Experiments}

\subsection{Experimental Setups}
\noindent \textbf{Datasets.} We totally utilize three popular real-time object detection datasets. \textbf{(1)} MS-COCO 2017~\cite{b15} has 143K images for three sub-sets. \textbf{(2) \& (3)} VOC~2007 \& 2012~\cite{b48} are parts of the Pascal VOC Challenges, totaling 20 object classes.  

\noindent \textbf{Implementations.} Our training is primarly consistent with the standard paractices~\cite{b1,b3}. We train models using 300 epochs and 3 epochs for warmup. We use SGD as the optimizer with 0.0005 weight decay and 0.937 SGD momentum. We adjust $lr$ using linear scaling with initial $lr$ 0.01 and the cosine schdule. The batch size is 256 on a machine with 8 $\times$ NVIDIA Tesla V100 (32GB). The FPS and latency are all tested with batchsize = 1 on a single V100.

\noindent \textbf{Evaluation Metrics.} We use different Average Precision (AP) as the evaluation metrics, which integrate precision and recall.

\subsection{Performance Comparisons}
We compare 3A-YOLO series detectors with representive CNN-based real-time detectors on COCO and VOC datasets. We totally deploy four resolutions for 3A-YOLO on COCO, enabling a comprehensive analysis of how the speed and accuracy of our models vary with different resolutions, as displayed in Tab.~\ref{tab:duibi-1}. The results demonstrate that our 3A-YOLO outperforms other methods in the speed-accuracy trade-off. Remarkably, 3A-YOLO-X exhibits 50.6\% AP, exceeding YOLOv4-CSP~\cite{b9} by 3.1\% AP with competitive 57.4 FPS. As illustrated in Tab.~\ref{tab:duibi-2}, we compare 3A-YOLO with other methods on Pascal VOC. We observe that although YOLOv5-L~\cite{b3} has a stronger backbone and larger input resolution, 3A-YOLO still achieves better performance, surpassing \cite{b3} by 3.7\% AP. Regarding 3A-YOLO-Tiny and Nano, they are also quite competitive compared to their counterparts, as displayed in Tab.~\ref{tab:duibi-3}. For example, our 3A-YOLO-Tiny increases 6.1\%AP on COCO compared to YOLOv4-Tiny~\cite{b9}.

\begin{table}[t!]
\caption{Comparisons of 3A-YOLO-Tiny and 3A-YOLO-Nano with their counterparts in terms of AP(\%) on MS-COCO~\emph{(val 2017)}.}
\centering
\vspace{-0.5em}
\setlength{\arrayrulewidth}{0.37pt}
\renewcommand{\arraystretch}{1.0}
\resizebox{\linewidth}{!}{
\begin{tabular}{lcccc}
\toprule
Method                & AP(\%) & Parameters & GFLOPs \\ \hline
PP-YOLO-Tiny~\cite{b23}                 & 22.7            & 4.20M               &     \textbf{--}            \\
YOLOv4-Tiny~\cite{b9}              & 19.8            & 6.05M               & 3.47            \\ 
NanoDet~\cite{b47}                 & 21.5            & 2.41M               & 1.88             \\ 
\rowcolor{cvprblue!20}
\textbf{3A-YOLO-Tiny (Ours)}              & \textbf{25.9}            & 6.17M               & 3.56            \\
\rowcolor{cvprblue!20}
\textbf{3A-YOLO-Nano (Ours)}               & \textbf{23.5}            & 3.42M               & 2.44            \\
\hline
\end{tabular}
}
\label{tab:duibi-3}
\vspace{-1.0em}
\end{table}

\begin{table}[t!]
\caption{Ablation studies of our improvements on MS-COCO~\emph{(val 2017)}.}
\centering
\vspace{-0.5em}
\setlength{\arrayrulewidth}{0.37pt}
\renewcommand{\arraystretch}{1.0}
\resizebox{\linewidth}{!}{
\begin{tabular}{lccccc}
\toprule
Method   & Parameters (M)  & AP(\%)     & FPS  \\ \hline
YOLOv4 Baseline    & 64.4   & 42.2       & 96.4 \\  

+ TDA-YOLO Module      & 65.2            & 44.0 \textcolor[rgb]{0.984,0.329,0.188}{(+1.8)} & 90.2 \\ 

+ Strong Data Augmentation     & 65.2           & 45.5 \textcolor[rgb]{0.984,0.329,0.188}{(+1.5)} & 90.2 \\  

+ Coordinate Attention (CA)        & 65.5      & 46.8 \textcolor[rgb]{0.984,0.329,0.188}{(+1.3)} & 89.2 \\  

+ Neck Network Improvements      & 65.1         & 47.7 \textcolor[rgb]{0.984,0.329,0.188}{(+0.9)} & 92.6 \\  

+ Tricks Deployment   & 65.1   & \textbf{48.2} \textcolor[rgb]{0.984,0.329,0.188}{(+0.5)} & 92.6 \\  

\hline
\textcolor[rgb]{0.501,0.541,0.529}{+ Dilated Encoder (optional)}  & \textcolor[rgb]{0.501,0.541,0.529}{67.8}   & \textcolor[rgb]{0.501,0.541,0.529}{48.6} \textcolor[rgb]{0.501,0.541,0.529}{(+0.4)} & \textcolor[rgb]{0.501,0.541,0.529}{81.3} \\  
\hline
\end{tabular}}
\label{tab:-ablation-overall}
\vspace{-1.1em}
\end{table}

\begin{table}[t]  
\caption{Ablation explorations on the effectiveness of three attention components in the TDA-YOLO Module on the PASCAL VOC dataset.}
\centering
\vspace{-0.5em}
\setlength{\arrayrulewidth}{0.37pt}
\renewcommand{\arraystretch}{1.0}
\resizebox{\linewidth}{!}{
\begin{tabular}{ccccc}
\toprule
Scale Att. & Spatial Att. & Task Att. & $\mathrm{AP_{50}}$(\%) & $\mathrm{AP_{75}}$(\%) \\ \hline
$\times$    & $\times$     & $\times$   & 85.7 & 66.9  \\
\checkmark   & $\times$       & $\times$  & 86.7 & 69.6 \\
\checkmark   & \checkmark   & $\times$    & 87.4 & 71.4 \\
\rowcolor{cvprblue!20}
\checkmark  & \checkmark  & \checkmark  & \textbf{87.9} & \textbf{71.7} \\
\hline
\end{tabular}}
\label{tab:three-att}
\end{table}

\begin{table}[] 
\caption{Performance comparisons about the effectiveness of intermediate attention components with different channel reduction ratios on the Pascal VOC dataset.}
\centering
\vspace{-0.2em}
\setlength{\arrayrulewidth}{0.37pt}
\renewcommand{\arraystretch}{1.02}
\resizebox{\linewidth}{!}{
\begin{tabular}{lcccc}
\toprule
Method      & Ratio & $\mathrm{AP_{50}}$(\%) & $\mathrm{AP_{75}}$(\%)\\ \hline

Previous 3A-YOLO      & \textbf{--}              & 85.7              & 66.9              \\
+ SE~\cite{b13}   & 16             & 86.7              & 67.5              \\
+ CBAM~\cite{b17} & 8              & 86.7              & 67.1              \\
+ ECA~\cite{b53}  & \textbf{--}              & 86.4              & 66.7              \\
+ CA~\cite{b4}    & 32             & 86.3              & 67.2              \\

\rowcolor{cvprblue!20}
+ CA~\cite{b4}    & 16             & \textbf{87.3}              & \textbf{67.6}              \\ \hline
\end{tabular}}
\label{tab:CA-ablations}
\vspace{-1.0em}
\end{table}

\subsection{Ablation Studies}
\noindent \textbf{Contributions of Improvement Measures.}~We progressively conduct in-depth ablation studies to explore the contributions of our improvement measures, as shown in Tab.~\ref{tab:-ablation-overall}. Note that the YOLOv4 baseline here only implement the network structure without extra tricks to perform clear comparisons. The results indicate that our measures yield explicit enhancements when juxtaposed against the baseline. Remarkably, our TDA-YOLO Module exhibits the largest gain of 1.8\% AP on COCO, demonstrating its superiority in unified discriminative learning. 

\noindent \textbf{Impact of Attention Components in our New Head.}~We further perform ablation explorations on the impact of three discriminative attention components in the TDA-YOLO Module on the Pascal VOC dataset, as illustrated in Tab.~\ref{tab:three-att}. We conclude that despite operating on different feature dimensions, three hierarchical components synergistically function as a unified attention block, collectively achieving triple discriminative awareness by complementing each other's strengths.

\noindent \textbf{Ablation Investigations on Coordinated Attention}
To further investigate the effectiveness of CA~\cite{b4} in coordinatedly capturing the inter-channel relationships and precise positional information, we compare CA against other advanced intermediate attention components under different channel reduction ratios, leveraging the previous 3A-YOLO as baseline on the Pascal VOC benchmark, as illustrated in Tab.~\ref{tab:CA-ablations}. We figure out that adding SE~\cite{b13} raises model performance by 1.0\% $\mathrm{AP_{50}}$. For CBAM~\cite{b17}, its spatial attention component seems not to contribute to 3A-YOLO compared to SE. The model performance of adding ECA~\cite{b53} increases by 0.7\% $\mathrm{AP_{50}}$, which is worse than SE~\cite{b53}. However, when we deploy the overall CA, the model performance remarkably increases by 1.6\% $\mathrm{AP_{50}}$ and 0.7\% $\mathrm{AP_{75}}$, respectively. Furthermore, we can clearly observe that increasing the channel reduction ratio from 16 to 32 is harmful to the intermediate feature fusion and representations learning.

\subsection{Visualization Results}
\begin{figure}[t!]
\centering
\includegraphics[width=\linewidth]{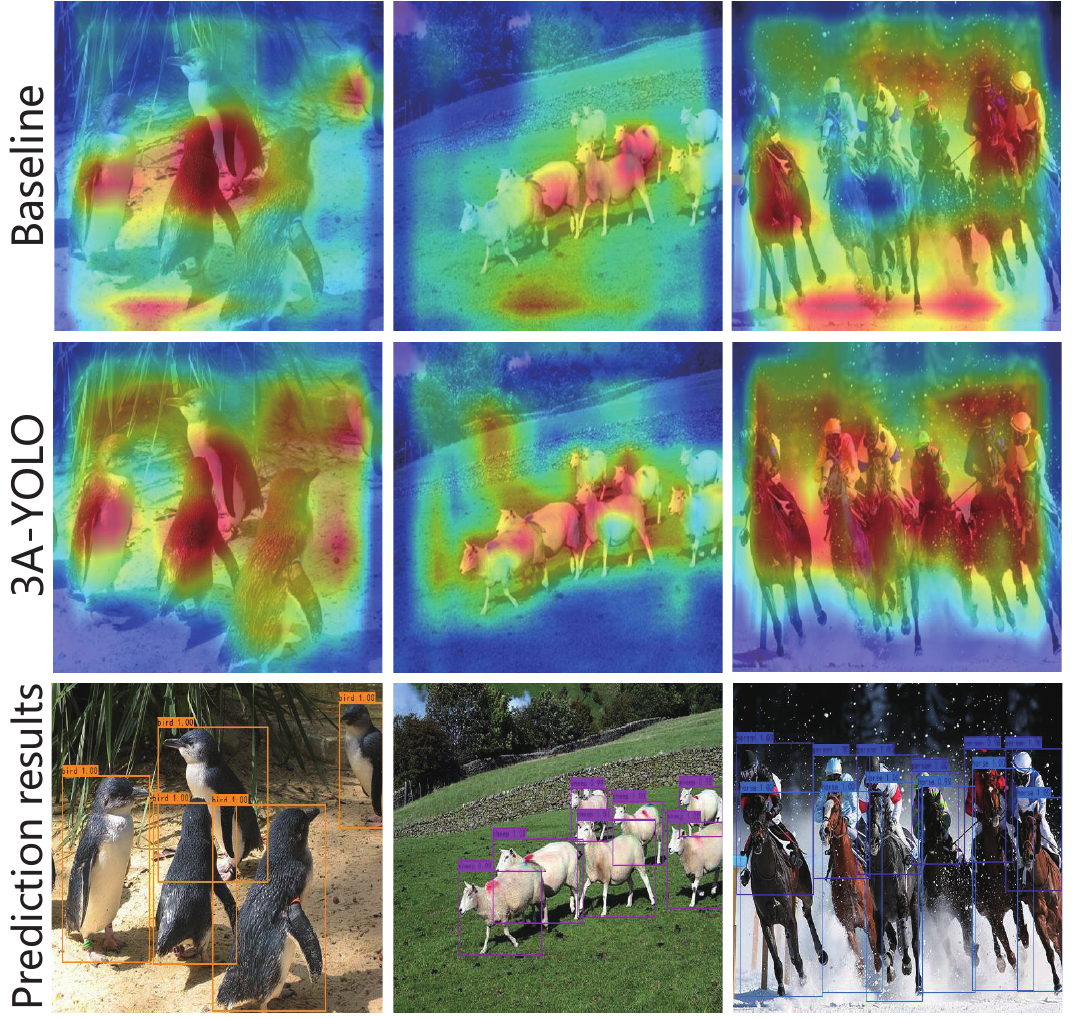}
\vspace{-1.5em}
\caption{Feature map visualizations coupled with the predictions of 3A-YOLO.}
\vspace{-1.0em}
\label{fig:visualizations}
\end{figure}

In Fig.~\ref{fig:visualizations}, we visualize the intermediate feature maps, output by the baseline and our 3A-YOLO. We further display the final predictions of our final detector. This visual evidence effectively validates the efficacy of our intermediate learning improvements measures for capturing and modeling the objects of interest, underscoring the capability of our 3A-YOLO detector to establish the precise location relationships between targeted areas.

\section{Conclusion and Discussions}
\label{sec:conclusion}
In this paper, a new series real-time detectors termed 3A-YOLO is presented. We introduce multiple attention to build the TDA-YOLO Module with triple discriminative awareness and complementarily learn the coordinated intermediate representations for better adaptability of our new head. The neck network improvements coupled with various tricks further enhance the robustness and generalization of our 3A-YOLO. Besides, we further develop three scaled model versions to meet diverse application needs. Extensive experimental results across two datasets verify the effectiveness of our detectors.

We leverage the simple yet effective baselines YOLOv4~\cite{b1} and YOLOv4-Tiny~\cite{b9} to evaluate the superiority of our improvement measures. However, the integration performance of our measures on more advanced baselines such as YOLOv7~\cite{YOLOv7} remains unexplored. We aim to address this research gap in our future developments.

\bibliographystyle{IEEEbib}
\bibliography{references}

\begin{thebibliography}{10}

\bibitem{b11}
Shaoqing Ren, Kaiming He, Ross Girshick, and Jian Sun,
\newblock ``Faster r-cnn: Towards real-time object detection with region proposal networks,''
\newblock {\em IEEE T-PAMI}, vol. 39, no. 6, pp. 1137--1149, 2016.

\bibitem{b3}
Jocher,
\newblock ``yolov5,''
\newblock {\em https://github.com/ultralytics/yolov5}.

\bibitem{b8}
Wei Liu, Dragomir Anguelov, Dumitru Erhan, Christian Szegedy, Scott Reed, Cheng-Yang Fu, and Alexander~C Berg,
\newblock ``Ssd: Single shot multibox detector,''
\newblock in {\em ECCV}. Springer, 2016, pp. 21--37.

\bibitem{b7}
Ali Farhadi and Joseph Redmon,
\newblock ``Yolov3: An incremental improvement,''
\newblock in {\em CVPR}, 2018, vol. 1804, pp. 1--6.

\bibitem{b12}
Mingxing Tan, Ruoming Pang, and Quoc~V Le,
\newblock ``Efficientdet: Scalable and efficient object detection,''
\newblock in {\em CVPR}, 2020, pp. 10781--10790.

\bibitem{b1}
Alexey Bochkovskiy, Chien-Yao Wang, and Hong-Yuan~Mark Liao,
\newblock ``Yolov4: Optimal speed and accuracy of object detection,''
\newblock {\em arXiv preprint arXiv:2004.10934}, 2020.

\bibitem{YOLOv7}
Chien-Yao Wang, Alexey Bochkovskiy, and Hong-Yuan~Mark Liao,
\newblock ``Yolov7: Trainable bag-of-freebies sets new state-of-the-art for real-time object detectors,''
\newblock in {\em CVPR}, 2023, pp. 7464--7475.

\bibitem{b13}
Jie Hu, Li~Shen, and Gang Sun,
\newblock ``Squeeze-and-excitation networks,''
\newblock in {\em CVPR}, 2018, pp. 7132--7141.

\bibitem{b53}
Qilong Wang, Banggu Wu, Pengfei Zhu, Peihua Li, Wangmeng Zuo, and Qinghua Hu,
\newblock ``Eca-net: Efficient channel attention for deep convolutional neural networks,''
\newblock in {\em CVPR}, 2020, pp. 11534--11542.

\bibitem{EBSE-YOLO}
Shuohe Wang, Yongda Wang, Yujian Chang, et~al.,
\newblock ``Ebse-yolo: high precision recognition algorithm for small target foreign object detection,''
\newblock {\em IEEE Access}, vol. 11, pp. 57951--57964, 2023.

\bibitem{YOLO-SE}
Tianyong Wu and Youkou Dong,
\newblock ``Yolo-se: Improved yolov8 for remote sensing object detection and recognition,''
\newblock {\em Applied Sciences}, vol. 13, no. 24, pp. 12977, 2023.

\bibitem{SBMYv3_Samal}
Sonali Samal, Yu-Dong Zhang, Thippa~Reddy Gadekallu, Rajashree Nayak, and Bunil~Kumar Balabantaray,
\newblock ``Sbmyv3: improved mobyolov3 a bam attention-based approach for obscene image and video detection,''
\newblock {\em Expert Systems}, vol. 40, no. 6, pp. e13230, 2023.

\bibitem{b14}
Jongchan Park, Sanghyun Woo, Joon-Young Lee, and In~So Kweon,
\newblock ``A simple and light-weight attention module for convolutional neural networks,''
\newblock {\em IJCV}, vol. 128, no. 4, pp. 783--798, 2020.

\bibitem{b17}
Sanghyun Woo, Jongchan Park, et~al.,
\newblock ``Cbam: Convolutional block attention module,''
\newblock in {\em ECCV}, 2018, pp. 3--19.

\bibitem{YOLO-SLD_Chung}
Ming-An Chung, Yu-Jou Lin, and Chia-Wei Lin,
\newblock ``Yolo-sld: An attention mechanism-improved yolo for license plate detection,''
\newblock {\em IEEE Access}, 2024.

\bibitem{b2}
Xiyang Dai, Yinpeng Chen, Bin Xiao, Dongdong Chen, Mengchen Liu, Lu~Yuan, and Lei Zhang,
\newblock ``Dynamic head: Unifying object detection heads with attentions,''
\newblock in {\em CVPR}, 2021, pp. 7373--7382.

\bibitem{b15}
Tsung-Yi Lin, Michael Maire, Serge Belongie, James Hays, Pietro Perona, Deva Ramanan, et~al.,
\newblock ``Microsoft coco: Common objects in context,''
\newblock in {\em ECCV}. Springer, 2014, pp. 740--755.

\bibitem{b48}
Mark Everingham, Luc Van~Gool, Christopher~KI Williams, John Winn, and Andrew Zisserman,
\newblock ``The pascal visual object classes (voc) challenge,''
\newblock {\em IJCV}, vol. 88, pp. 303--338, 2010.

\bibitem{b16}
T-YLPG Ross and GKHP Doll{\'a}r,
\newblock ``Focal loss for dense object detection,''
\newblock in {\em CVPR}, 2017, pp. 2980--2988.

\bibitem{b28}
Kaiming He, Xiangyu Zhang, Shaoqing Ren, and Jian Sun,
\newblock ``Deep residual learning for image recognition,''
\newblock in {\em CVPR}, 2016, pp. 770--778.

\bibitem{b30}
Chien-Yao Wang, Hong-Yuan~Mark Liao, Yueh-Hua Wu, Ping-Yang Chen, Jun-Wei Hsieh, and I-Hau Yeh,
\newblock ``Cspnet: A new backbone that can enhance learning capability of cnn,''
\newblock in {\em CVPR workshops}, 2020, pp. 390--391.

\bibitem{b58}
Seung-Wook Kim, Hyong-Keun Kook, Jee-Young Sun, Mun-Cheon Kang, and Sung-Jea Ko,
\newblock ``Parallel feature pyramid network for object detection,''
\newblock in {\em ECCV}, 2018, pp. 234--250.

\bibitem{b36}
Shu Liu, Lu~Qi, Haifang Qin, Jianping Shi, and Jiaya Jia,
\newblock ``Path aggregation network for instance segmentation,''
\newblock in {\em CVPR}, 2018, pp. 8759--8768.

\bibitem{b4}
Qibin Hou, Daquan Zhou, and Jiashi Feng,
\newblock ``Coordinate attention for efficient mobile network design,''
\newblock in {\em CVPR}, 2021, pp. 13713--13722.

\bibitem{b49}
Chaofeng Chen, Dihong Gong, Hao Wang, Zhifeng Li, and Kwan-Yee~K Wong,
\newblock ``Learning spatial attention for face super-resolution,''
\newblock {\em IEEE TIP}, vol. 30, pp. 1219--1231, 2020.

\bibitem{b31}
Jie Hu, Li~Shen, Samuel Albanie, Gang Sun, and Andrea Vedaldi,
\newblock ``Gather-excite: Exploiting feature context in convolutional neural networks,''
\newblock {\em NeurIPS}, vol. 31, 2018.

\bibitem{b34}
Drew Linsley, Dan Shiebler, Sven Eberhardt, and Thomas Serre,
\newblock ``Learning what and where to attend,''
\newblock {\em arXiv preprint arXiv:1805.08819}, 2018.

\bibitem{b41}
Kaiming He, Xiangyu Zhang, Shaoqing Ren, and Jian Sun,
\newblock ``Spatial pyramid pooling in deep convolutional networks for visual recognition,''
\newblock {\em IEEE T-PAMI}, vol. 37, no. 9, pp. 1904--1916, 2015.

\bibitem{b9}
Chien-Yao Wang, Alexey Bochkovskiy, and Hong-Yuan~Mark Liao,
\newblock ``Scaled-yolov4: Scaling cross stage partial network,''
\newblock in {\em CVPR}, 2021, pp. 13029--13038.

\bibitem{b39}
Yinpeng Chen, Xiyang Dai, Mengchen Liu, Dongdong Chen, et~al.,
\newblock ``Dynamic relu,''
\newblock in {\em ECCV}. Springer, 2020, pp. 351--367.

\bibitem{b26}
Zhaohui Zheng, Ping Wang, Wei Liu, Jinze Li, Rongguang Ye, and Dongwei Ren,
\newblock ``Distance-iou loss: Faster and better learning for bounding box regression,''
\newblock in {\em AAAI}, 2020, vol.~34, pp. 12993--13000.

\bibitem{b46}
Qiang Chen, Yingming Wang, Tong Yang, Xiangyu Zhang, Jian Cheng, and Jian Sun,
\newblock ``You only look one-level feature,''
\newblock in {\em CVPR}, 2021, pp. 13039--13048.

\bibitem{b23}
Xiang Long et~al.,
\newblock ``Pp-yolo: An effective and efficient implementation of object detector,''
\newblock {\em arXiv preprint arXiv:2007.12099}, 2020.

\bibitem{b37}
Songtao Liu, Di~Huang, and Yunhong Wang,
\newblock ``Learning spatial fusion for single-shot object detection,''
\newblock {\em arXiv preprint arXiv:1911.09516}, 2019.

\bibitem{PP-YOLOv2}
Xin Huang, Xinxin Wang, Wenyu Lv, Xiaying Bai, Xiang Long, Kaipeng Deng, Qingqing Dang, Shumin Han, Qiwen Liu, Xiaoguang Hu, et~al.,
\newblock ``Pp-yolov2: A practical object detector,''
\newblock {\em arXiv preprint arXiv:2104.10419}, 2021.

\bibitem{b59}
Jiale Cao, Yanwei Pang, Jungong Han, and Xuelong Li,
\newblock ``Hierarchical shot detector,''
\newblock in {\em ICCV}, 2019, pp. 9705--9714.

\bibitem{b50}
Nikita Dvornik, Konstantin Shmelkov, Julien Mairal, and Cordelia Schmid,
\newblock ``Blitznet: A real-time deep network for scene understanding,''
\newblock in {\em ICCV}, 2017, pp. 4154--4162.

\bibitem{b51}
Kyungyul Kim, ByeongMoon Ji, Doyoung Yoon, and Sangheum Hwang,
\newblock ``Self-knowledge distillation with progressive refinement of targets,''
\newblock in {\em ICCV}, 2021, pp. 6567--6576.

\bibitem{b52}
Yousong Zhu, Chaoyang Zhao, Jinqiao Wang, Xu~Zhao, Yi~Wu, and Hanqing Lu,
\newblock ``Couplenet: Coupling global structure with local parts for object detection,''
\newblock in {\em ICCV}, 2017, pp. 4126--4134.

\bibitem{b47}
Rangi Lyu,
\newblock ``Nanodet-plus: Super fast and high accuracy lightweight anchor-free object detection model,''
\newblock {\em https://github. com/RangiLyu/nanodet}, 2021.

\end{thebibliography}

\end{document}